\documentclass[11pt]{article}

\usepackage[top=0.75in, bottom=0.75in, left=1in, right=1in]{geometry}

\makeatletter
\renewcommand\section{\@startsection{section}{1}{\z@}%
  {-0.24in}{0.10in}{\large\bf\raggedright}}
\renewcommand\subsection{\@startsection{subsection}{2}{\z@}%
  {-0.20in}{0.08in}{\normalsize\bf\raggedright}}
\renewcommand\subsubsection{\@startsection{subsubsection}{3}{\z@}%
  {-0.18in}{0.08in}{\normalsize\sc\raggedright}}
\makeatother

\usepackage[utf8]{inputenc}
\usepackage[T1]{fontenc}

\usepackage{amsmath}
\usepackage{amssymb}
\usepackage{booktabs}
\usepackage{multirow}
\usepackage{float}

\usepackage{tikz}
\usetikzlibrary{positioning, arrows.meta, fit, calc, shapes.geometric}

\usepackage{listings}
\usepackage{microtype}
\usepackage{xcolor}
\usepackage{needspace}

\usepackage{natbib}
\usepackage[colorlinks,citecolor=blue!70!black,urlcolor=blue!70!black,linkcolor=black]{hyperref}
\usepackage{url}

\newcommand{\pkg}{\textsc{Causal-TS}}
\newcommand{\grace}{\textsc{Grace}}
\newcommand{\reap}{\textsc{Reap}}
\newcommand{\lucid}{\textsc{Lucid}}
\newcommand{\cdnots}{\textsc{CDNOTS}}
\newcommand{\cdnotsplus}{\textsc{CDNOTS+}}
\newcommand{\sypi}{\textsc{Cedar}}

\title{Causal-TS: A Python Library for Causal Discovery\\in Stationary and Nonstationary Time Series}

\author{\textbf{Mohammad Fesanghary}  \texttt{mfesanghary1@bloomberg.net} \\
  Bloomberg LP, New York, NY, USA
}

\date{}

\begin{document}

\maketitle


\begin{abstract}
We describe \pkg{}, an open-source Python library that provides a common
interface and lagged-graph representation for causal discovery (structure
learning) from stationary and nonstationary multivariate time series. It
learns lagged and contemporaneous causal graphs with constraint-based (\cdnots{}, \cdnotsplus{}, \sypi{}), score-based (LGES,
TGES, GES, LASSO-VAR, VARLiNGAM, Granger) and hybrid (\grace{}, \reap{})
methods. The constraint-based methods share 17 linear and nonlinear
conditional independence (CI) tests, nine with PyTorch implementations.
It also provides candidate-regime detection with per-regime discovery,
latent-confounder correction (\lucid{}),
missing- and mixed-data support, a data-driven configuration recommender, a
command-line interface, and an optional DoWhy bridge from a discovered graph
to causal effect estimates and counterfactuals. The library is
pip-installable, tested on Python 3.10--3.14, and available at
\texttt{https://github.com/bloomberg/causal-ts}.
\end{abstract}

\noindent\textbf{Keywords:} Causal Discovery, Structure Learning, Time Series, Nonstationarity,
  Conditional Independence, GPU Acceleration, Python

\section{Introduction}
\label{sec:intro}

Multivariate time series are not merely i.i.d.\ tables with timestamps.
Autocorrelation, lagged and contemporaneous relations, and changing mechanisms
complicate causal discovery---but can also provide identifying structure.
Temporal order excludes future-to-past lagged edges, while nonstationarity can
resolve directions left ambiguous in stationary data
\citep{huang2020causal,sadeghi2025cdnots}. These problems arise across climate
science, neuroscience, finance, epidemiology, and
economics~\citep{peters2017elements}.

Causal-learn~\citep{zheng2024causal} and
Tigramite~\citep{runge2019detecting} provide mature, widely used foundations
for causal discovery: causal-learn offers broad constraint- and score-based
methods, while Tigramite provides a rich framework for temporal discovery,
regimes, latent confounding, and effect estimation.
DoWhy~\citep{sharma2020dowhy} supports downstream causal inference. \pkg{} is
designed to complement and interoperate with these ecosystems. It places
constraint-based, score-based, and hybrid temporal methods behind one interface
and graph format, uses nonstationarity through a time-index node, and provides
17 CI tests (nine with PyTorch implementations) and a data-driven configuration
recommender. Its CI tests run inside Tigramite's PCMCI family, its GPU tests
register with causal-learn's PC, FCI, and CD-NOD, and discovered graphs connect
to DoWhy for effect estimation and counterfactuals.

\section{Interface and Usage}
\label{sec:interface}

A typical \pkg{} workflow has three steps (Figure~\ref{fig:arch}):
load data, run discovery, and analyze results.

\begin{figure}[t]
\centering
\begin{tikzpicture}[scale=0.55, every node/.style={transform shape},
    >=Stealth,
    box/.style={draw, rounded corners=3pt, minimum height=0.7cm,
                font=\small, inner sep=6pt, fill=#1},
    box/.default=white,
    arr/.style={->, thick, color=black!70},
    grp/.style={draw, rounded corners=4pt, dashed, inner sep=2.5pt,
                color=black!40},
]

\node[box=blue!6, minimum width=2.4cm, minimum height=2.8cm] (data) at (-1.0,-0.675) {};
\node[font=\normalsize\bfseries, above=2pt of data.north] {Time series};
\draw[blue!70,   line width=1pt] plot[smooth,tension=0.6]
  coordinates{(-1.75,0.10)(-1.40,0.30)(-1.05,0.05)(-0.70,0.25)(-0.35,0.00)(-0.25,0.15)};
\draw[red!60,    line width=1pt] plot[smooth,tension=0.6]
  coordinates{(-1.75,-0.37)(-1.40,-0.22)(-1.05,-0.42)(-0.70,-0.25)(-0.35,-0.45)(-0.25,-0.30)};
\draw[green!55!black,line width=1pt] plot[smooth,tension=0.6]
  coordinates{(-1.75,-0.80)(-1.40,-0.62)(-1.05,-0.85)(-0.70,-0.65)(-0.35,-0.88)(-0.25,-0.72)};
\draw[orange!75!black,line width=1pt] plot[smooth,tension=0.6]
  coordinates{(-1.75,-1.22)(-1.40,-1.40)(-1.05,-1.20)(-0.70,-1.42)(-0.35,-1.22)(-0.25,-1.38)};

\node[box=yellow!15, minimum width=3.8cm, align=center, font=\scriptsize]
  (ci) at (4.5, 2.5)
  {\textbf{\normalsize 17 CI Tests (CPU / GPU)}\\[2pt]
   {\small ParCorr\,\textbullet\,KCI\,\textbullet\,SplitKCI\,\textbullet\,RCOT\,\textbullet\,DFCIT}\\
   {\small RAFT\,\textbullet\,GCM\,\textbullet\,PCM\,\textbullet\,CMIknn\,\textbullet\,GCMI\,\textbullet\,\dots}};

\node[box=green!12, text width=5.7cm, align=center, minimum height=0.65cm]
  (cdnots)     at (4.5, 0.60)  {\cdnots{}\,/\,\cdnotsplus{}\,/\,\sypi{}};
\node[box=green!12, text width=5.7cm, align=center, minimum height=0.65cm]
  (cdnotsplus) at (4.5,-0.25)  {LGES\,/\,TGES\,/\,GES};
\node[box=green!12, text width=5.7cm, align=center, minimum height=0.65cm]
  (sypi)       at (4.5,-1.10)  {\footnotesize\mbox{LASSO-VAR\,/\,VARLiNGAM\,/\,Granger}};
\node[box=green!12, text width=5.7cm, align=center, minimum height=0.65cm]
  (grace)      at (4.5,-1.95)  {\grace{}\,/\,\reap{} refinement};

\node[grp, fit=(cdnots)(cdnotsplus)(sypi)(grace)] (algogrp) {};

\node[font=\small, color=black!50, below=8pt of algogrp.south] (baselines)
  {+ \lucid{} confounder correction\,\textbullet\,regime discovery};

\draw[arr] (ci.south) -- (algogrp.north);

\node[box=red!8, text width=3.2cm, minimum height=0.9cm, align=center]
  (graph)  at (10.6,  0.30) {Causal graph\\+ visualization};
\node[box=red!8, text width=3.2cm, minimum height=0.9cm, align=center]
  (effect) at (10.6, -1.65) {Effects, SCMs,\\counterfactuals (DoWhy)};

\draw[arr] (data.east) -- (algogrp.west);
\draw[arr] (algogrp.east) -- (graph.west);
\draw[arr] (algogrp.east) -- (effect.west);

\node[font=\small, color=black!45, below=4pt of baselines.south]
  {\texttt{causal-ts discover data.csv {-}{-}algorithm cdnots {-}{-}max-lag 3}};

\end{tikzpicture}
\caption{\pkg{} architecture. Constraint-based methods share the CI tests,
which \grace{} and \reap{} use for their default candidate. The CLI and the
Python API expose every method.}
\label{fig:arch}
\end{figure}

\needspace{12\baselineskip}
\paragraph{Python API.}
Discovery takes one call per method:
\begin{lstlisting}
from causalts import inspect_df, discover_df, run_reap
from causalts.lges import lges_discovery

rec = inspect_df(df)["recommendation"]  # algorithm, CI test, lag, C preset
res = discover_df(df, algorithm=rec["algorithm"], ci_test=rec["ci_test"],
                  max_lag=rec["max_lag"], include_C=rec["include_C"],
                  c_preset=rec["c_preset"])
lucid = res.deconfound()                # latent-confounder correction
reap = run_reap(df, max_lag=3)          # refines its own CDNOTS candidate
G, info = lges_discovery(df, max_lag=3) # score-based (BIC), plain arrays
\end{lstlisting}
Result objects plot themselves and bridge to DoWhy via
\texttt{.estimate\_effect()}, \texttt{.fit\_scm()}, and \texttt{.counterfactual()};
plain graph arrays go through \texttt{wrap\_graph}.
Domain knowledge enters through \texttt{Ancestral\-Knowledge}, which accepts
direct edge constraints (enforced during skeleton discovery) and path-level
ancestral constraints (applied during orientation; unmet ones are reported).

\paragraph{Command-line interface.}
The \texttt{causal-ts} CLI covers the same workflow:
\begin{lstlisting}[language=bash]
causal-ts inspect data.csv     # data health + recommended configuration
causal-ts discover data.csv --algorithm cdnots --max-lag 3
\end{lstlisting}
\texttt{causal-ts install-skill} installs a packaged Claude Code/Codex agent
skill for inspection, cost-gated discovery, and graph explanation;
\texttt{discover {-}{-}json} returns a machine-readable summary.

\section{Package Design}
\label{sec:design}

\subsection{Algorithms}

\pkg{} exposes constraint-based, hybrid, and score-based methods through a
shared lagged-graph representation (Table~\ref{tab:algos}).

\begin{table}[t]
\centering
\caption{Discovery methods in \pkg{}.}
\label{tab:algos}
\small
\begin{tabular}{@{}lll@{}}
\toprule
\textbf{Method} & \textbf{Family} & \textbf{Key mechanism} \\
\midrule
\cdnots~\citep{sadeghi2025cdnots}  & Constraint (PC)     & Nonstationarity through $C$ \\
\cdnotsplus{}                      & Constraint (PCMCI+) & Parent-conditioned MCI \\
\sypi~\citep{fesanghary2025cedar}  & Pairwise CI         & AR-aware lag discovery \\
\grace~\citep{fesanghary2025grace} & Hybrid ($L_0$ gates) & Gated candidate refinement \\
\reap{}                            & Hybrid (regression) & Spline $F$-test refinement \\
LGES / TGES / GES                  & Score (BIC)         & Lag-embedded equivalence search \\
LASSO-VAR / VARLiNGAM / Granger    & Score (VAR fit)     & Linear baselines \\
\bottomrule
\end{tabular}
\end{table}

\cdnots{} extends PC with temporal constraints and a potentially
multidimensional variable $C$---trend bases, user-supplied seasonal bases,
or regime indicators---that captures nonstationarity.
\cdnotsplus{} replaces the skeleton phase with a PCMCI+-style
procedure~\citep{runge2020discovering} that conditions only on discovered
parents, limiting conditioning-set growth.
\sypi{} uses partial distance correlation lag selection, autoregression-aware CI
testing, and MCI pruning to avoid exponential conditioning sets.
\grace{} and \reap{} refine the lagged edges of any directed candidate graph
(\cdnots{} by default) by grow--shrink rounds: \grace{} prunes with Hard Concrete
gates~\citep{louizos2018learning}, \reap{} by partial $F$-tests on additive
spline regressions.
The score-based methods search Markov equivalence classes on lag-embedded
data with edges into the past forbidden: LGES~\citep{ejaz2025lges} inserts
edges less greedily than GES~\citep{chickering2002optimal}, and
TGES~\citep{larsen2025tges} orients edges by time tiers.
VARLiNGAM and Granger wrap causal-learn's implementations; GES adapts the
search code of LGES and forbids edges into the past; LASSO-VAR is native.

\paragraph{Confounders, regimes, and data.}
\lucid{}~\citep{fesanghary2026lucid} routes from the VAR-residual spectrum
to a sparse or a pervasive (factor) confounding branch and applies the
matching correction to a compatible discovered graph.
The \texttt{causalts.regime} module detects or accepts candidate regimes---via
PELT~\citep{killick2012optimal}, HMM, CUSUM, KS, Levene, quantile, seasonal,
or ensemble detectors---and runs discovery within each regime.
\cdnots{}, \cdnotsplus{}, and \sypi{} handle missing values by pairwise
masking or VAR-EM imputation, and discrete columns by stratified,
discrete-aware tests.

\subsection{Conditional Independence Tests and GPU Acceleration}

The constraint-based methods share 17 CI tests (19 implementations, counting
CPU and GPU versions) spanning linear (ParCorr), kernel (KCI,
SplitKCI~\citep{pogodin2024practical}, LinSig, inspired by
\citealp{manten2024signature}), random-feature (RCOT, RAFT), residual-score
(GCM~\citep{shah2020hardness}, PCM), distribution-free
(DFCIT~\citep{cai2022distribution}), nearest-neighbour
(CMIknn~\citep{runge2018conditional}), mixed-data, and discrete families.
Nine have PyTorch implementations for CUDA or CPU; custom tests plug in
through a registry.

\subsection{Software Quality}

\pkg{} installs via \texttt{pip install causalts} (extras \texttt{[dowhy]},
\texttt{[tigramite]}, \texttt{[parquet]}, \texttt{[dev]}). \textbf{Testing:}
80 pytest modules on Python 3.10--3.14, with GitHub Actions for tests, linting,
documentation, and OpenSSF Scorecard. \textbf{Documentation:} Sphinx API
reference, 32 example notebooks and a tutorial, a CI test selection guide, and
a CLI reference. \textbf{Dependencies \& license:} core dependencies include
NumPy, Pandas, PyTorch, SciPy, NetworkX, causal-learn, and statsmodels;
GPL-3.0-or-later (paper CC-BY 4.0).

\section{Interoperability and Related Software}
\label{sec:comparison}

Table~\ref{tab:compare} summarizes documented functionality, not statistical
accuracy or runtime. The libraries are complementary: \pkg{} reuses
causal-learn's VARLiNGAM and Granger implementations, lets its CI
tests run inside Tigramite and causal-learn discovery workflows, and delegates
optional effect estimation and counterfactual analysis to DoWhy.

\begin{table}[t]
\centering
\caption{Feature comparison with related Python libraries (documented public
APIs, checked November 2026). \checkmark\ native; $\circ$ restricted, optional,
or bridge-based; -- absent.}
\label{tab:compare}
\small
\begin{tabular}{lcccc}
\toprule
\textbf{Feature} & \textbf{\pkg{}} & \textbf{causal-learn} & \textbf{Tigramite} & \textbf{DoWhy} \\
\midrule
Temporal graph discovery            & \checkmark & $\circ$$^a$ & \checkmark & -- \\
Nonstationarity-aware discovery     & \checkmark & $\circ$$^b$ & \checkmark$^c$ & -- \\
Score-based graph search            & \checkmark & \checkmark  & --          & -- \\
Latent-confounding support          & $\circ$$^d$ & \checkmark$^e$ & \checkmark$^f$ & -- \\
GPU-accelerated CI tests            & \checkmark & --          & $\circ$$^g$ & -- \\
Nonlinear CI tests for discovery    & \checkmark & \checkmark & \checkmark & -- \\
Effect estimation                   & $\circ$$^h$ & --         & \checkmark & \checkmark \\
Regime-aware discovery              & \checkmark$^i$ & --          & \checkmark  & -- \\
\bottomrule
\multicolumn{5}{@{}p{0.95\linewidth}@{}}{\scriptsize $^a$\,Granger and VARLiNGAM. $^b$\,CD-NOD, not time-series specific.
$^c$\,RPCMCI, J-PCMCI+. $^d$\,restricted post-discovery correction (\lucid{}). $^e$\,FCI. $^f$\,LPCMCI.
$^g$\,GPDCtorch only. $^h$\,optional DoWhy bridge. $^i$\,candidate-regime detection or supplied labels, then per-regime discovery.}\\
\end{tabular}
\end{table}

\section{Conclusion}
\label{sec:conclusion}

\pkg{} provides a common workflow for constraint-based, score-based, and
hybrid causal discovery in stationary and nonstationary time series, with
integrations for causal-learn, Tigramite, and DoWhy. Its guarantees remain
those of the selected method and data regime: it assumes sufficiently sampled
series, \lucid{} supports only its documented sparse and pervasive-factor
branches, and selection bias is outside the current scope. The submitted
version is v0.28.1 (\url{https://github.com/bloomberg/causal-ts}), archived
on Zenodo at \href{https://doi.org/10.5281/zenodo.24205661}{doi:10.5281/zenodo.24205661}.

\section*{Acknowledgments}
We thank the open-source contributors to Tigramite, causal-learn, and
the GES and LGES implementations, whose code informed or is adapted in parts
of \pkg{} (see \texttt{NOTICE} for attribution).

\vspace{-4pt}
\bibliographystyle{plainnat}
\bibliography{references}

\end{document}